\documentclass{article}
\usepackage{arxiv}
\usepackage{amsmath}
\usepackage{wrapfig}
\usepackage{graphicx}
\usepackage{subcaption}

\title{Behavior Structformer: Learning Players Representations with Structured 
Tokenization}

\author{
    Oleg Smirnov\thanks{Equal contribution.}\\
	\texttt{oleg.smirnov@king.com}\\
    \And
    Labinot Polisi\footnotemark[1]\\
	\texttt{labinot.polisi@king.com} \\
 }

\begin{document}
\date{}
\maketitle

\begin{abstract}
    In this paper, we introduce the Behavior Structformer, a method for modeling user behavior using structured tokenization within a Transformer-based architecture. By converting tracking events into dense tokens, this approach enhances model training efficiency and effectiveness. We demonstrate its superior performance through ablation studies and benchmarking against traditional tabular and semi-structured baselines. The results indicate that structured tokenization with sequential processing significantly improves behavior modeling.
\end{abstract}

\section{Introduction}
The landmark Transformer~\cite{vaswani2017attention} model has demonstrated 
impressive performance across a wide range of scenarios, extending well beyond the 
realm of Natural Language Processing. The potential of multi-head self-attention 
method lies in the ability to pick up a signal from any data modality, provided it 
exhibits a spatial (e.g., sequential) structure and is appropriately preprocessed 
into discrete tokens for model consumption. However, in practice, the convergence 
rate of Transformer models in default configurations is considered unsatisfactory.
This issue can be mitigated by incorporating prior domain knowledge and inductive 
biases during the tokenization phase, making the data more suitable for processing
by the algorithm.

In the field of Computer Vision, the Hybrid Vision Transformers 
approach~\cite{dosovitskiy2020image} has shown that leveraging a pre-trained
convolutional backbone as a feature extractor leads to faster convergence and 
improved downstream performance. Similar observations have been made in customer 
modeling for personalization~\cite{luo2023mcm}, where purchase and non-purchase 
actions were pre-embedded before processing with a BERT-like model. In the healthcare 
sector, a sequence of electronic health records was preprocessed based on domain
expert knowledge to be further consumed by a Transformer-based model with an
objective to predict the next medical code~\cite{li2020behrt}.

Inspired by these advances, we propose a method for modeling in-game player 
behavior data that employs a structured approach to convert tracking events into 
dense tokens. We benchmark and compare the proposed approach against the tabular
and semi-structured baselines.

%% TODO: Research Questions

\section{Related Work}
Structured behavior data is common and widespread in many practical settings,
where users interact with systems and select from a predefined set of
possible actions. Their interactions lead to \emph{events} that are described by
metadata fields and are typically recorded in system logs. Possible applications
span various industries and sectors, ranging from e-commerce 
stores~\cite{sun2019bert4rec,luo2023mcm} and retail 
banking~\cite{zhang2018sequential,zhu2020modeling}, to 
healthcare~\cite{li2020behrt} and online gaming~\cite{vuorre2023intensive}.

Contrary to time-series data, which depends on continuously sampled functions, 
structured behavior data is usually not represented this way. Instead, individual 
behavioral events may include values that are ordinal, categorical, or derived
from more complex functions. Those collections of values, often denoted as
\emph{features} or \emph{variables}, may vary between different events and do not 
always align. This variability can thwart the modeling efforts, as it requires 
capturing signals  under sparsity while also accounting for the interactions between 
functions and their temporal characteristics.

Previous studies have investigated the use of domain-specific transformations to 
encode individual features into vectorized embeddings~\cite{zhang2018sequential}.
These embeddings are then processed simultaneously by a Recurrent Neural Network
to capture the sequential modality, and a Convolutional Neural Network to encode
the interrelationships between activities. While the proposed architecture 
outperforms a baseline k-Nearest Neighbor classifier in an online fraud detection 
task, its complexity restricts its practical applicability.

Another work explored a hierarchical approach in which numerical and categorical 
features for each event are initially embedded and processed using a field-level 
extractor to encodes interactions up to the second order~\cite{zhu2020modeling}.
Next, an event-level extractor is employed to capture higher-order interactions. 
While designed with explainability in mind, which is crucial for the fraud
detection domain, this method lacks sufficient flexibility to learn from 
long-range contextual information.

In the e-commerce domain, BERT4Rec~\cite{sun2019bert4rec} and the Multi-task
Customer  Model~\cite{luo2023mcm} methods focus on adapting a BERT-based
multi-head self-attention architecture for recommender systems and personalization
tasks. Due to the uniformity of online shopping data, all events can be represented 
using the same set of features. This allows to simplify the encoding process to
just embedding and average pooling steps.

A parallel line of work introduced a method for converting sequences of structured
behavior events into flat, string-based formats to utilize pre-trained language 
models for mining information directly from textual data~\cite{wang2024player2vec}. 
However, this transformation results in a format with low entropy and constrained 
vocabulary, which causes subpar learning performance under moderate dataset sizes.

\section{Methodology}
\subsection{In-game behavior data}
Player behavior data, though superficially similar to tracking data in other
fields, exhibits several distinct challenges that can complicate its modeling.
Firstly, video games are crafted to be dynamic and engaging, resulting in user
interactions occurring more frequently than in Web browsing or other applications.
Secondly, in-game events vary significantly in terms of their metadata, that
can be also attributed to the rich and diverse gaming environment. In a hypothetical
scenario of behavior data collected from an online casual game, possible events
might include starting a game, selecting a particular difficulty level, and
interacting with various in-game content, such as commercials. However, among
these event types, the overlap in feature sets may be minimal, including only a 
numerical timestamp field and a categorical player identifier.

In this work, we use a dataset of player behavior sessions collected from a large 
mobile game provider over 30 days with $1\text{M}$ players uniformly 
sampled from the user base. The resulting database consists of
$\approx 3\text{M}$ sessions, where $90\%$ is allocated for the training split
and the rest for the validation split.
\subsection{Structured tokenization}

\begin{figure}[t]
\centering
\includegraphics[clip, trim=2.7cm 0cm 0cm 0cm, width=0.8\textwidth]{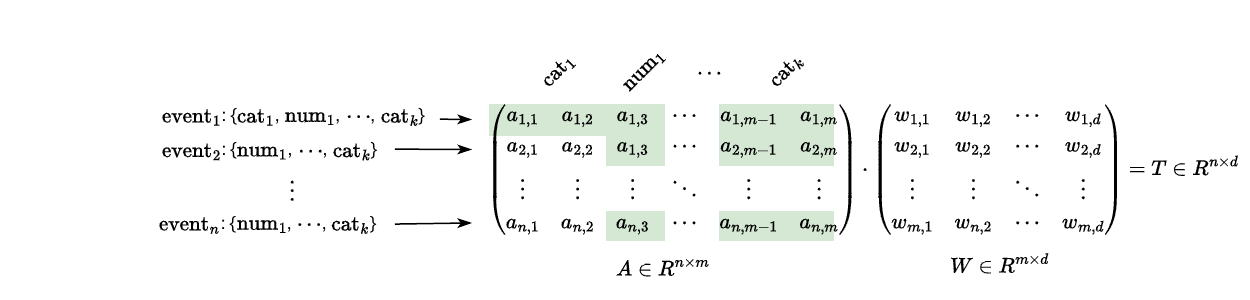}
\caption{Sparse structured tokenization method. Green elements of the feature
matrix $A$ denote non-zero entries corresponding to non-missing features of input
events. For example, the last event in sequence $\text{event}_n$ does not include
the first categorical feature $\text{cat}_1$, which leads to the matrix entries
$a_{n,1}$ and $a_{n, 2}$ having zero values.}
\label{fig:token_sparse}
\end{figure}

We represent a time-ordered sequence of events as a series of dense vectors, which
are then processed as tokens within a Transformer-based architecture. This approach
addresses the issue of feature variability among events through a sparse structured tokenization, as depicted in Figure~\ref{fig:token_sparse}. Specifically, in an input
sequence of $n$ events, each $\text{event}_i$, $i \in 1, 2, \ldots, n$ is
determined by a collection of numerical $\text{num}_j$ and categorical
$\text{cat}_j$ features, where $j \in 1, 2, \ldots, k$,
$k = k_{\text{cat}} + k_{\text{num}}$ is the total number of supported features,
and $k_{\text{cat}}$ and $k_{\text{num}}$ are the numbers of distinct categorical
and numerical features correspondingly. During encoding, categorical features are
mapped to dense vectors as rows of look-up embedding matrices $M^{(j)}$, whereas
numerical features are represented directly as one-dimensional vectors of their values:
\begin{equation*}
    v_j =
    \begin{cases}
    M^{(j)}_{\text{cat}_j} & \in R^c \\
    \text{num}_j & \in R
    \end{cases}
\end{equation*}
Where $c$ is the embeddings dimensionality hyperparameter. In instances where an
event is missing certain categorical or numerical features, these are represented
by zero vectors with the corresponding number of dimensions. The resulting sequence
of embedding vectors is concatenated to form a row of the feature matrix $A$:
\begin{equation*}
    A_j = [v_1, v_2, \ldots, v_k] \in R^m
\end{equation*}
Where $m = c k_{\text{cat}} + k_{\text{num}}$. The final matrix
$A \in R^{n \times m}$ representing the whole sequence is projected with a linear
layer $W \in R^{m \times d}$ to capture high-order interactions and produce
$d$-dimensional tokens.

The preprocessing pipeline retains a subset of event features which are known
to be informative for player representation learning based on the domain
expertise.

\subsection{Modeling objective}
The primary research objective of behavior modeling is to identify patterns that 
possess scientific or business value. However, in the absence of sufficient
ground-truth annotations, simpler proxies informed by domain knowledge in the 
relevant area can serve as indicators of the downstream performance.

\begin{figure}[ht]
\centering
\includegraphics[width=0.5\textwidth]{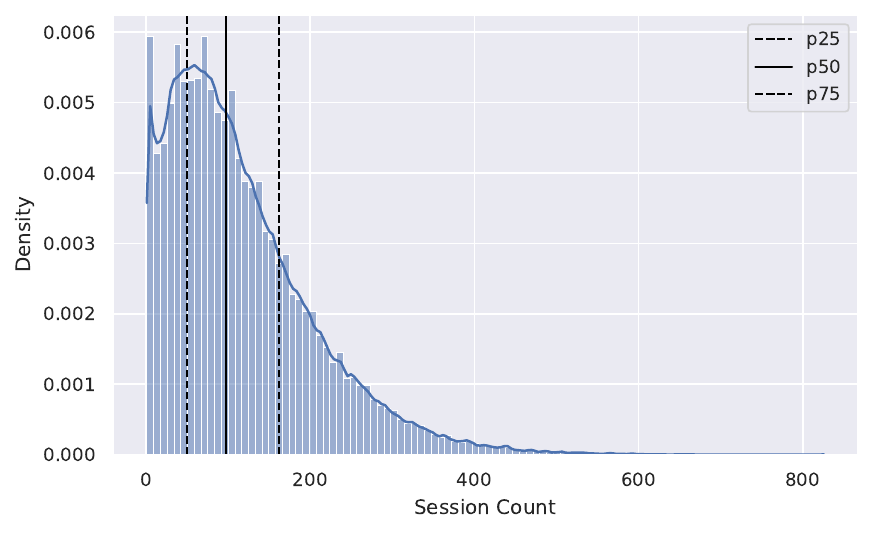}
\caption{Session count distribution along with the quantile values.}
\label{fig:session_count_distribution}
\end{figure}

\begin{figure}[h!]
    \centering
    \begin{subfigure}[t]{0.45\textwidth}
        \centering
        \includegraphics[height=4cm]{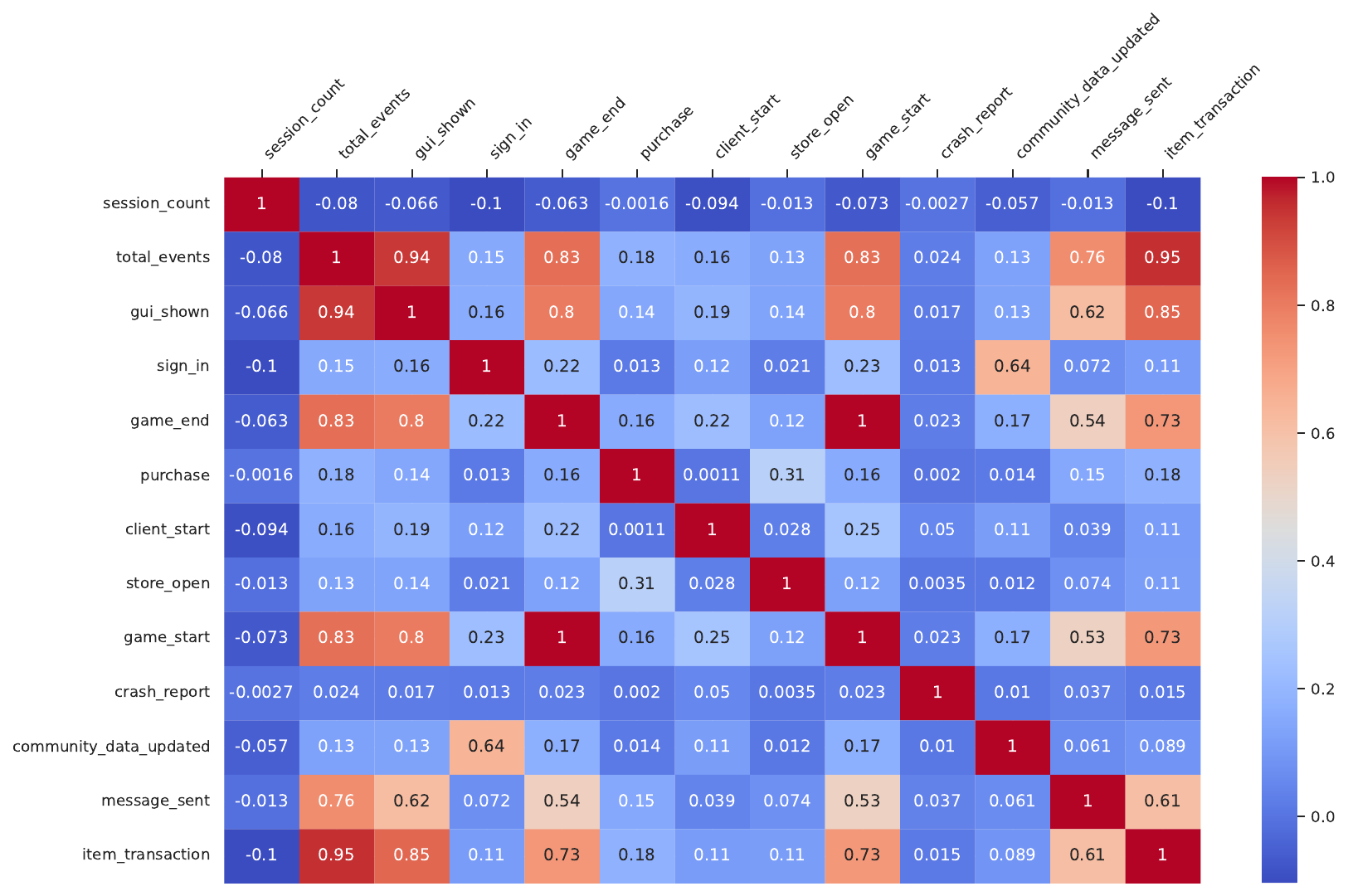}
        \caption{Correlation matrix showing the relationship between the number
        of sessions (target variable) and the counts of various event types.}
    \end{subfigure}%
    ~ 
    \begin{subfigure}[t]{0.45\textwidth}
        \centering
        \includegraphics[height=3.5cm]{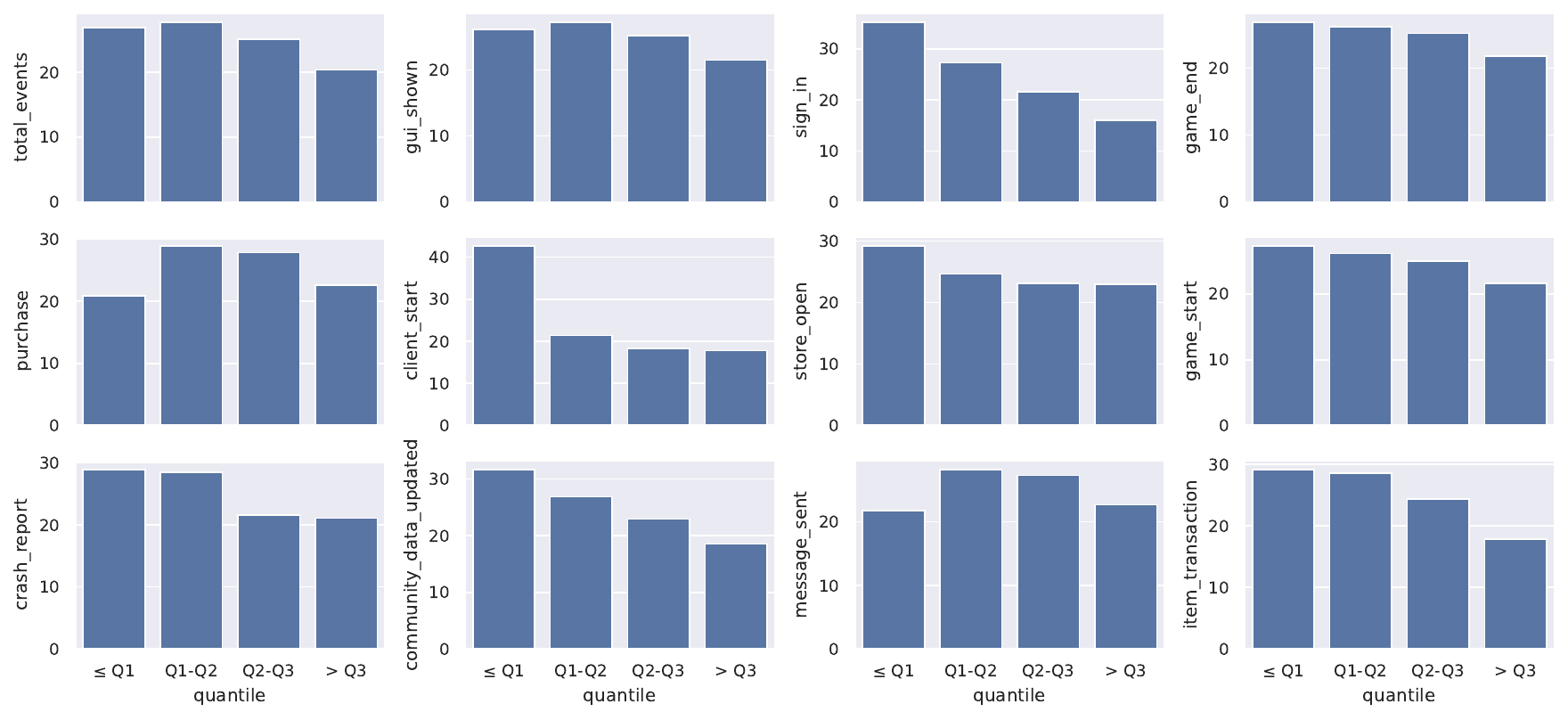}
        \caption{Percentages of average counts of various event types per the session
        number quantiles.}
    \end{subfigure}
    \caption{Player behavior dataset statistics.}
    \label{fig:dataset_statistics}
\end{figure}

We hypothesize that a single in-game behavior session holds sufficient information
to predict whether a user will play devotedly in the future or only occasionally.
The distribution of session counts per user is illustrated in 
Figure~\ref{fig:session_count_distribution}. As expected, this distribution 
exhibits a long-tailed shape, suggesting significant variations in player engagement 
levels. Furthermore, the exploratory analysis of the dataset illustrated in 
Figure~\ref{fig:dataset_statistics} reveals a correlation between player 
engagement and session content, indirectly supporting the hypothesis.
Hence, we fit the distribution with three alternative training objectives 
that progressively increase in difficulty: i) binary classification, which predicts 
whether a user's session count will be above or below the median, ii) 
multi-class classification, with classes defined by the ranges between quartiles;
and iii) regression, aimed at directly predicting the number of sessions.
By construction, the first two objectives work with the balanced class
distributions.

%% \subsection{Self-supervised pre-training}
%% TODO: masked event modeling motivated by masked image modeling and masked language modeling
%% TODO: directly using pixel-level auto-encoding (i.e., recovering the pixels of masked patches) for vision pre-training pushes the model to focus on short-range dependencies and high-frequency details~\cite{ramesh2021zero}
%% TODO: predicting spans rather than individual events as a more suitable objective for behavior data~\cite{joshi2020spanbert}

\section{Experiments}

\subsection{Backbone network}
A series of encoded event tokes is fed into a model backbone to learn
short-range and long-range inter-token dependencies. To this end, a
Transformer~\cite{vaswani2017attention} architecture is employed, where an
input sequence is prepended with a learnable \texttt{[CLS]} token that serves
as an aggregate representation for the entire sequence. To evaluate the necessity
of multi-headed self-attention in modeling behavior data, we investigate an 
alternative design that utilizes a Multi-Layer Perceptron (MLP) backbone. In this
configuration, input event tokens are aggregated with average pooling before 
processing with hidden linear layers, and no special classification token
is used.

We conduct experimental benchmarks on various hyperparameter settings, as detailed
in Table~\ref{tab:model_backbones}, to evaluate the capacity and generalization
capabilities of the proposed models. For this purpose, MLP model depths and widths
were chosen to roughly align with the Transformer model sizes. Both backbone
variants make use of the same number of embedding dimension, $c=32$, in each 
categorical feature embeddings layer.

In all experiments, the models were trained using the AdamW 
optimizer~\cite{loshchilov2018decoupled} with no weight decay and a batch
size of $128$. The learning rate was managed via a reduce on plateau scheduler,
starting at an initial value of $10^{-4}$, a decay factor of $10^{-1}$, with
the minimum threshold set at $10^{-6}$. 

\begin{table}[ht]
    \begin{tabular}{l c c c c}
        Backbone & Hidden dims & Layers & Heads & Params\\
        \hline\hline
        Structformer-Tiny   &  64 & 1 & 1 & 3M \\
        Structformer-Small  &  64 & 4 & 4 & 4M \\
        Structformer-Medium & 512 & 1 & 1 & 7M \\
        Structformer-Large  & 512 & 4 & 4 & 20M \\
        Structformer-XLarge & 512 & 8 & 8 & 37M \\
    \end{tabular}
    \quad
    \begin{tabular}{l c c}
        Backbone & Neurons & Params \\
        \hline\hline
        MLP-Small  & 256,128,64 & 3M \\
        MLP-Medium & 512,256,128 & 3.3M \\
        MLP-Large  & 512,512,256,128 & 3.5M \\
    \end{tabular}
    \caption{Model hyperparameters for Transformer- and MLP-based
    backbone architectures.}
    \label{tab:model_backbones}
\end{table}

\subsection{Tabular baseline}
We further perform an ablation study to assess whether fine-grained 
structured behavior data modeling is essential for the task at hand. For
this purpose, the dataset was converted into a count-based tabular format,
with each column representing the frequency of a corresponding event type 
per session, thus fully stripping the underlying sequential structure and
also omitting per-event features. The resulting tabular data was used to 
train a gradient boosting LightGBM predictor~\cite{ke2017lightgbm} with the 
same modeling objectives, where the hyperparameters were tuned to reach a 
strong baseline.

\subsection{Results}
We present standard classification metrics for both binary and multi-class
cases computed on the validation split in 
Table~\ref{tab:engagement_metrics_binary} and
Table~\ref{tab:engagement_metrics_multi}, respectively. The mean and standard 
deviation values were calculated from three independent runs using different
seeds.

\begin{table}[ht]
    \centering
    \begin{tabular}{l c c c c c c}
        Model & Accuracy~$\uparrow$ & F$_1$~$\uparrow$ & Precision~$\uparrow$ & Recall~$\uparrow$ & Train loss~$\downarrow$ & Val loss~$\downarrow$\\
        \hline\hline
        LightGBM & $.641 \pm .0$ & $.641 \pm .0$ & $.642 \pm .0$ & $.641 \pm .0$ & $.620 \pm .0$ & $.630 \pm .0$ \\
        \hline
        MLP-Small  & $.716 \pm .0$ & $.714 \pm .0$ & $.713 \pm .0$ & $.713 \pm .0$ & $.540 \pm .0$ & $.550 \pm .0$ \\
        MLP-Medium & $.715 \pm .0$ & $.713 \pm .0$ & $.713 \pm .0$ & $.712 \pm .0$ & $.533 \pm .0$ & $.551 \pm .0$ \\
        MLP-Large  & $.715 \pm .0$ & $.713 \pm .0$ & $.713 \pm .0$ & $.713 \pm .0$ & $.531 \pm .0$ & $.550 \pm .0$ \\
        \hline
        Structformer-Tiny   & $.722 \pm .0$ & $.722 \pm .0$ & $.723 \pm .0$ & $.722 \pm .0$ & $.535 \pm .0$ & $.537 \pm .0$ \\
        Structformer-Small  & $\mathbf{.731 \pm .0}$ & $\mathbf{.731 \pm .0}$ & $\mathbf{.731 \pm .0}$ & $\mathbf{.731 \pm .0}$ & $.515 \pm .0$ & $.526 \pm .0$ \\
        Structformer-Medium & $.726 \pm .0$ & $.726 \pm .0$ & $.727 \pm .0$ & $.726 \pm .0$ & $.525 \pm .0$ & $.531 \pm .0$ \\
        Structformer-Large  & $.728 \pm .0$ & $.728 \pm .0$ & $.729 \pm .0$ & $.728 \pm .0$ & $.522 \pm .0$ & $.529 \pm .0$ \\
        Structformer-XLarge & $.621 \pm .1$ & $.614 \pm .1$ & $.630 \pm .1$ & $.621 \pm .1$ & $.693 \pm .0$ & $.634 \pm .0$ \\
    \end{tabular}
    \caption{Binary classification performance metrics for determining
    whether a player's session count exceeds the dataset median.}
    \label{tab:engagement_metrics_binary}
\end{table}

\begin{table}[ht]
    \centering
    \begin{tabular}{l c c c c c c}
        Model & Accuracy~$\uparrow$ & F$_1$~$\uparrow$ & Precision~$\uparrow$ & Recall~$\uparrow$ & Train loss~$\downarrow$ & Val loss~$\downarrow$\\
        \hline\hline
        LightGBM & $.383 \pm .0$ & $.368 \pm .0$ & $.369 \pm .0$ & $.382 \pm .0$ & $1.276 \pm .0$ & $1.285 \pm .0$ \\
        \hline
        MLP-Small  & $.435 \pm .0$ & $.453 \pm .0$ & $.450 \pm .0$ & $.436 \pm .0$ & $1.158 \pm .0$ & $1.166 \pm .0$ \\
        MLP-Medium & $.434 \pm .0$ & $.451 \pm .0$ & $.448 \pm .0$ & $.435 \pm .0$ & $1.149 \pm .0$ & $1.167 \pm .0$ \\
        MLP-Large  & $.437 \pm .0$ & $.452 \pm .0$ & $.449 \pm .0$ & $.438 \pm .0$ & $1.144 \pm .0$ & $1.166 \pm .0$ \\
        \hline
        Structformer-Tiny   & $.468 \pm .0$ & $.456 \pm .0$ & $.455 \pm .0$ & $.465 \pm .0$ & $1.134 \pm .0$ & $1.135 \pm .0$ \\
        Structformer-Small  & $.478 \pm .0$ & $.470 \pm .0$ & $.469 \pm .0$ & $.476 \pm .0$ & $1.107 \pm .0$ & $1.118 \pm .0$ \\
        Structformer-Medium & $.470 \pm .0$ & $.459 \pm .0$ & $.457 \pm .0$ & $.468 \pm .0$ & $1.129 \pm .0$ & $1.131 \pm .0$ \\
        Structformer-Large  & $\mathbf{.480 \pm .0}$ & $\mathbf{.472 \pm .0}$ & $\mathbf{.470 \pm .0}$ & $\mathbf{.478 \pm .0}$ & $1.099 \pm .0$ & $1.113 \pm .0$ \\
        Structformer-XLarge & $.469 \pm .0$ & $.460 \pm .0$ & $.458 \pm .0$ & $.467 \pm .0$ & $1.129 \pm .0$ & $1.132 \pm .0$ \\
    \end{tabular}
    \caption{Multi-class classification performance metrics for determining
    player session counts across quantile ranges.}
    \label{tab:engagement_metrics_multi}
\end{table}

In our experiments, the final variance of the metrics is consistently
close to zero in almost all cases, suggesting low epistemic uncertainty and robust 
model convergence. The top-performing model in the binary classification scenario 
is Structformer-Small. Remarkably, further increases in model capacity, either through 
expanding the width (number of hidden dimensions) or depth (number of Transformer 
layers and self-attention heads), do not always yield significant improvement in the
metrics. Furthermore, all models reached a loss plateau after $\approx 250\text{K}$
steps midway through the training process. We theorize that the dataset comprising of
$1\text{M}$ users, may not contain sufficient signal to leverage higher-capacity
models in a simple binary classification case effectively.

The largest model evaluated, Structformer-XLarge, experienced an early collapse
during training. This issue could stem from problems with initialization or the 
learning rate schedule, and can be addressed with advanced training procedures
as typically required for large Transformer-based 
architectures~\cite{devlin2019bert}. We leave exploring solutions to these
problems for future work.

In the context of multi-class quantile prediction, we observe similar patterns
with a notable distinction that the Structformer-Large variant surpasses other 
models in performance. This leads to the conclusion that the optimal model
capacity is case-dependent and cannot simply be determined by choosing the
largest model available. In both tasks, the best-performing model and the
second-best model feature four internal layers, suggesting that depth may have
a more significant impact than token dimensionality.

Crucially, the proposed structured tokenization methodology in combination
with Transformer-based sequence modeling enhances downstream performance,
as compared to GBM- and MLP-based alternatives. Detailed plots of the 
training and validation curves, along with correposponding metrics, can be found
in Appendix~\ref{sec:detailed_curves}.

\section{Conclusion and Future Work}
In this study, we present preliminary results from modeling structured behavioral
data using a sparsity-aware encoding methodology, integrated with a self-attention 
mechanism. We demonstrate efficacy of the proposed approach on a synthetic task 
designed to predict player engagement, that involves fitting the statistics related
to session distribution. Our method consistently outperforms tabular-based baselines 
and simpler algorithms that do not take the sequential nature of behavior data into
account.

Informed by these findings, in future work we will focus on benchmarking the 
structured tokenization method on more realistic downstream applications and
comparing it to alternative approaches that use language modeling. Additionally,
we aim to explore contrastive learning and other self-supervised frameworks
to fully leverage large-scale data corpora. Finally, we plan to utilize a 
model explainability library~\cite{kokhlikyan2020captum} to gain insights into 
the informativeness of various behavioral events and their features across 
different tasks.

\bibliographystyle{abbrv}
\bibliography{main}

\clearpage
\appendix
\section{Metrics and Loss Curves}
\label{sec:detailed_curves}

\begin{figure}[ht]
\centering
\includegraphics[width=0.8\textwidth]{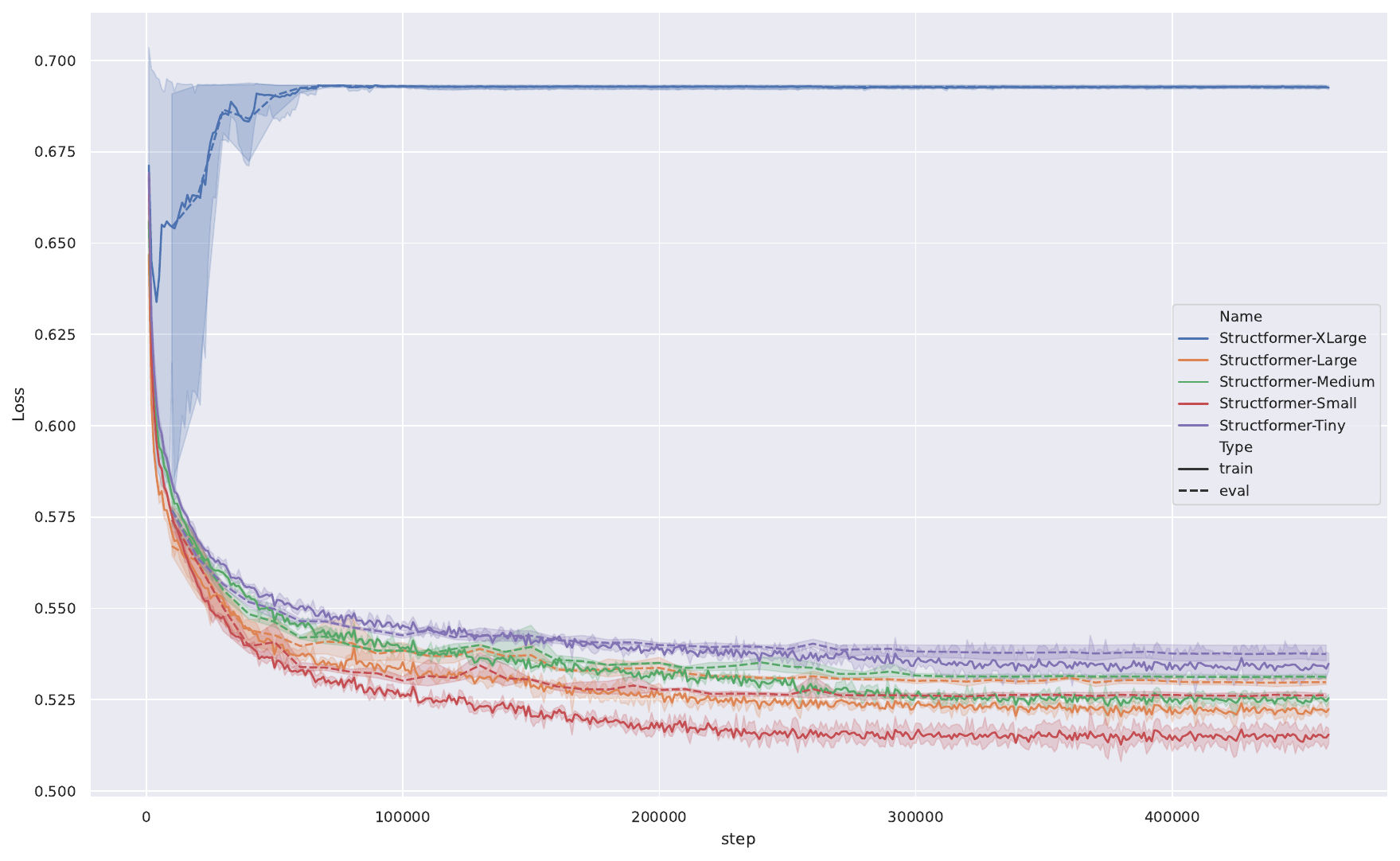}
\caption{Structformer binary classification loss curves.}
\label{fig:binary_loss_curves_combined}
\end{figure}

\begin{figure}[ht]
\centering
\includegraphics[width=0.8\textwidth]{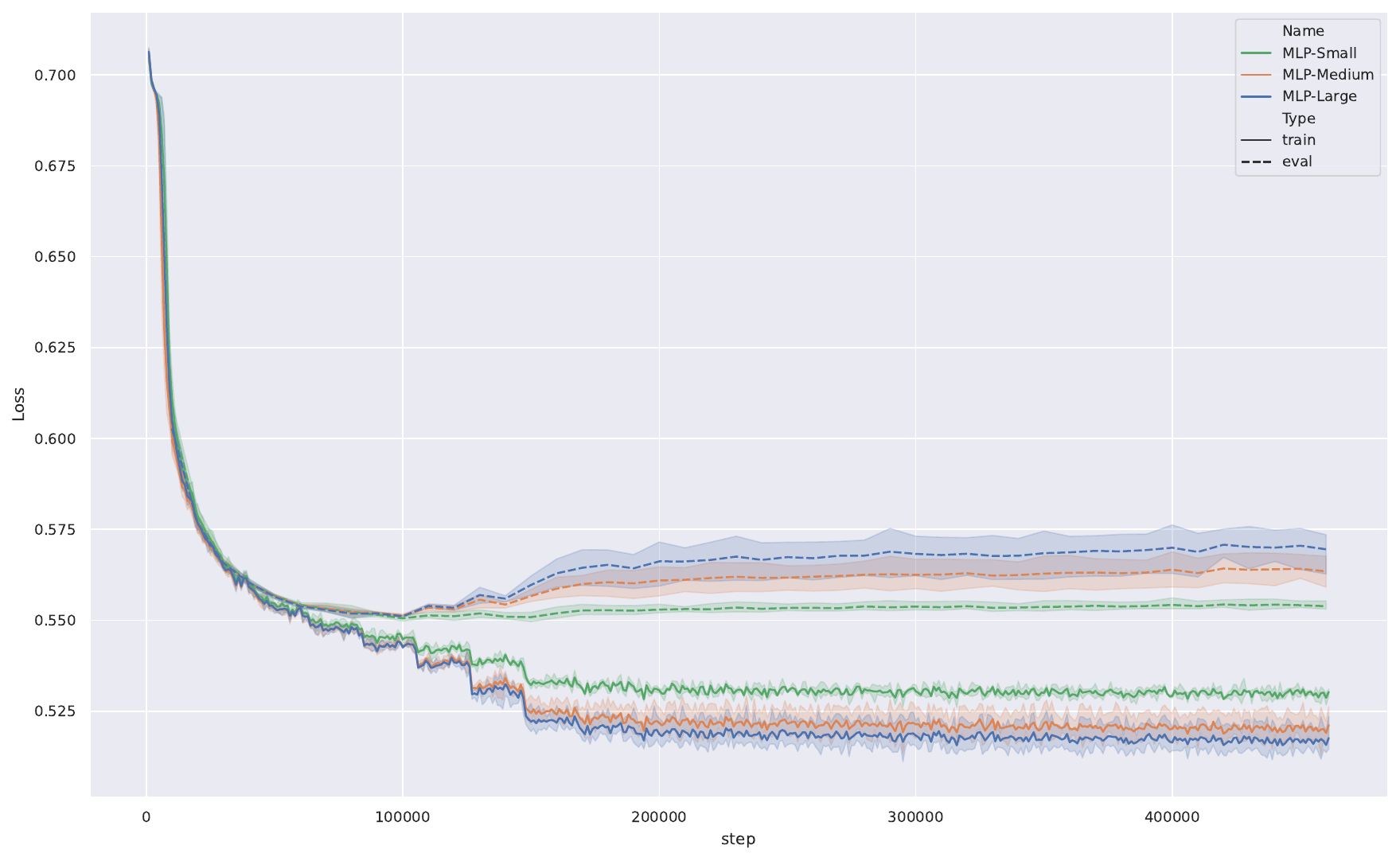}
\caption{MLP binary classification loss curves.}
\label{fig:mlp_binary_loss_curves_combined}
\end{figure}

\begin{figure}[ht]
\centering
\includegraphics[width=0.8\textwidth]{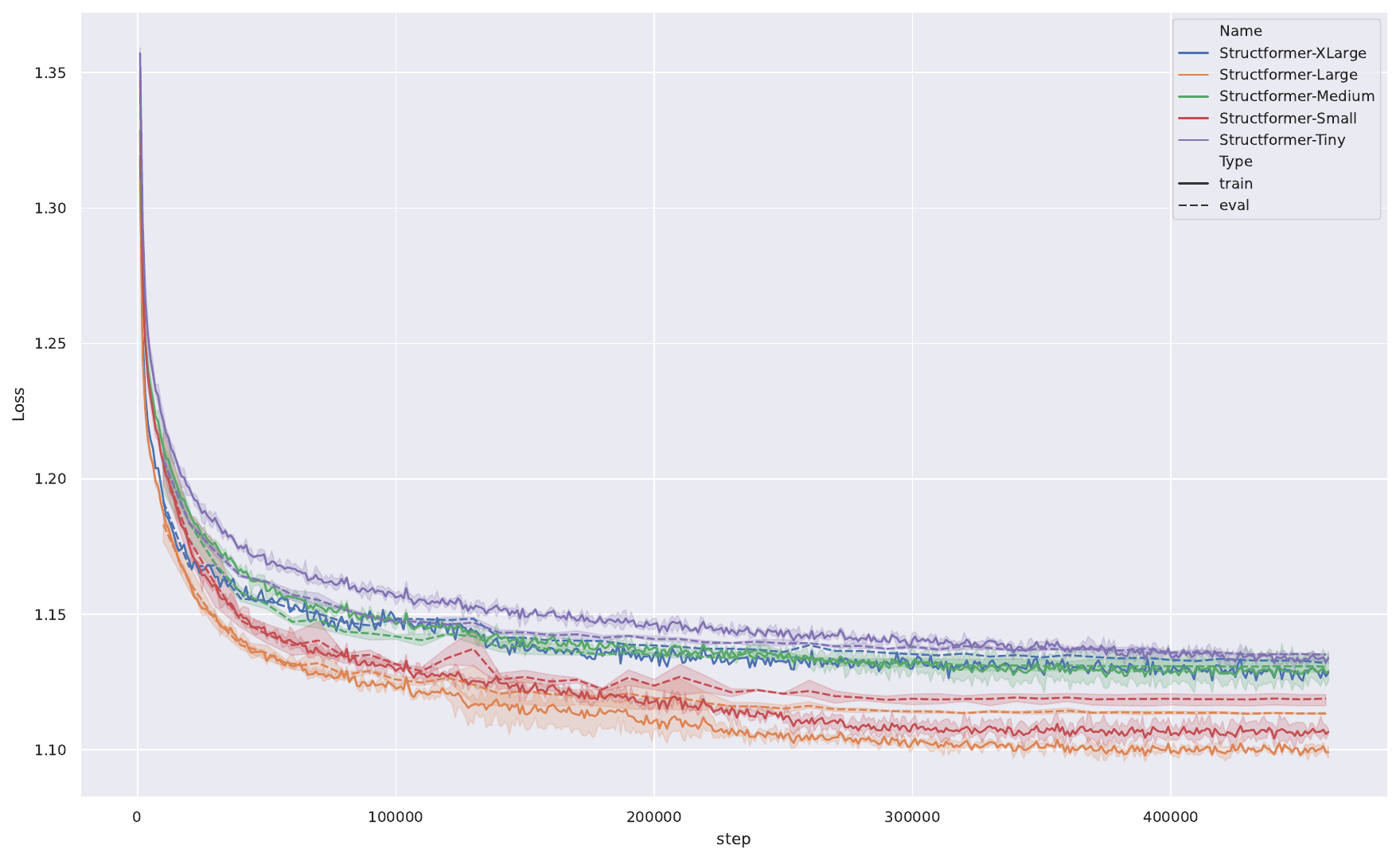}
\caption{Structformer multi-class classification loss curves.}
\label{fig:multiclass_loss_curves_combined.pdf}
\end{figure}

\begin{figure}[ht]
\centering
\includegraphics[width=0.8\textwidth]{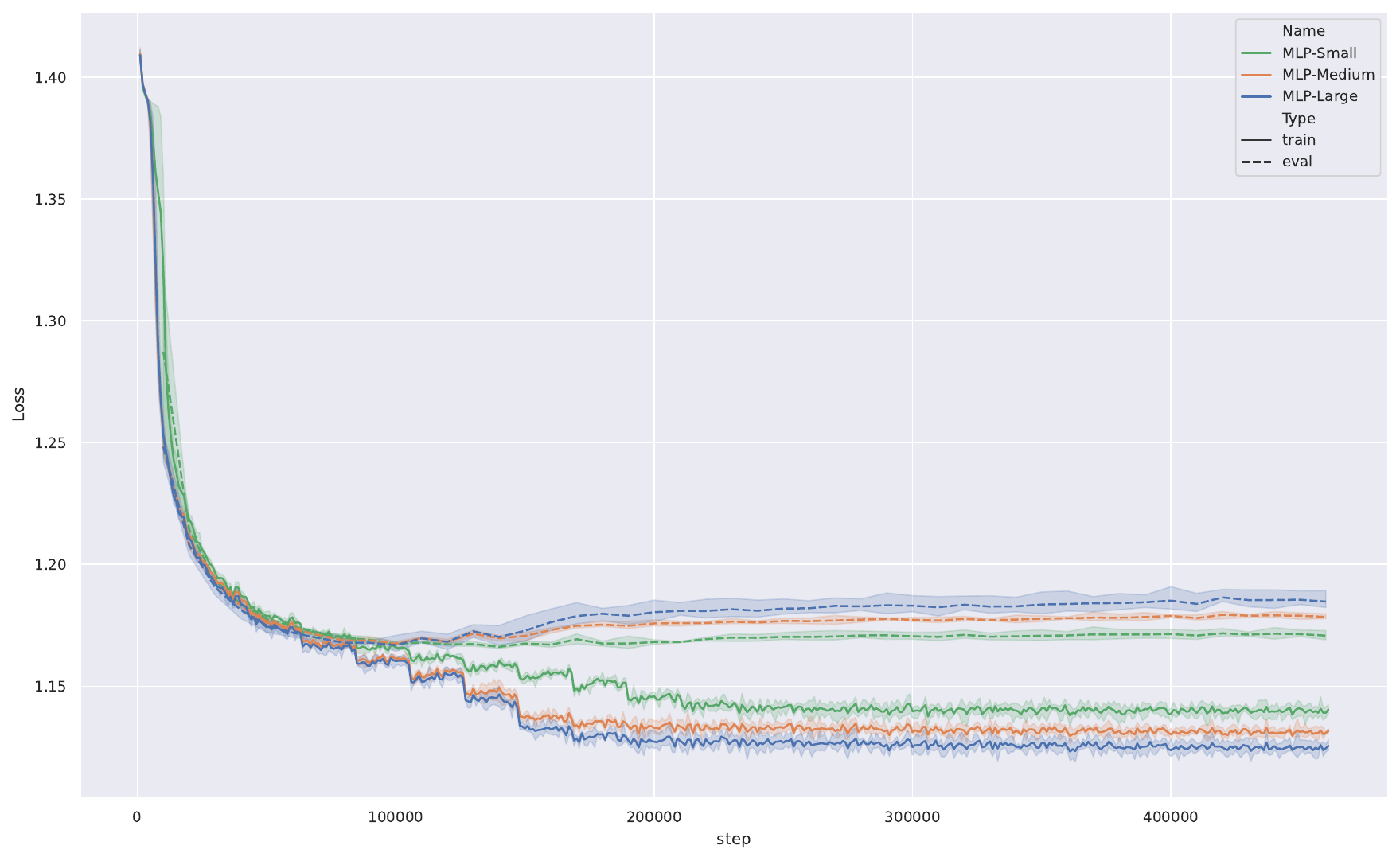}
\caption{MLP multi-class classification loss curves.}
\label{fig:mlp_multiclass_loss_curves_combined.pdf}
\end{figure}

% \begin{figure}[ht]
% \centering
% \includegraphics[width=0.5\textwidth]{binary_loss_curves.pdf}
% \caption{Binary classification loss curves.}
% \label{fig:binary_loss_curves}
% \end{figure}

% \begin{figure}[ht]
% \centering
% \includegraphics[width=0.5\textwidth]{multiclass_loss_curves.pdf}
% \caption{Multi-class classification loss curves.}
% \label{fig:multiclass_loss_curves}
% \end{figure}

\begin{figure}[ht]
\centering
\includegraphics[width=\textwidth]{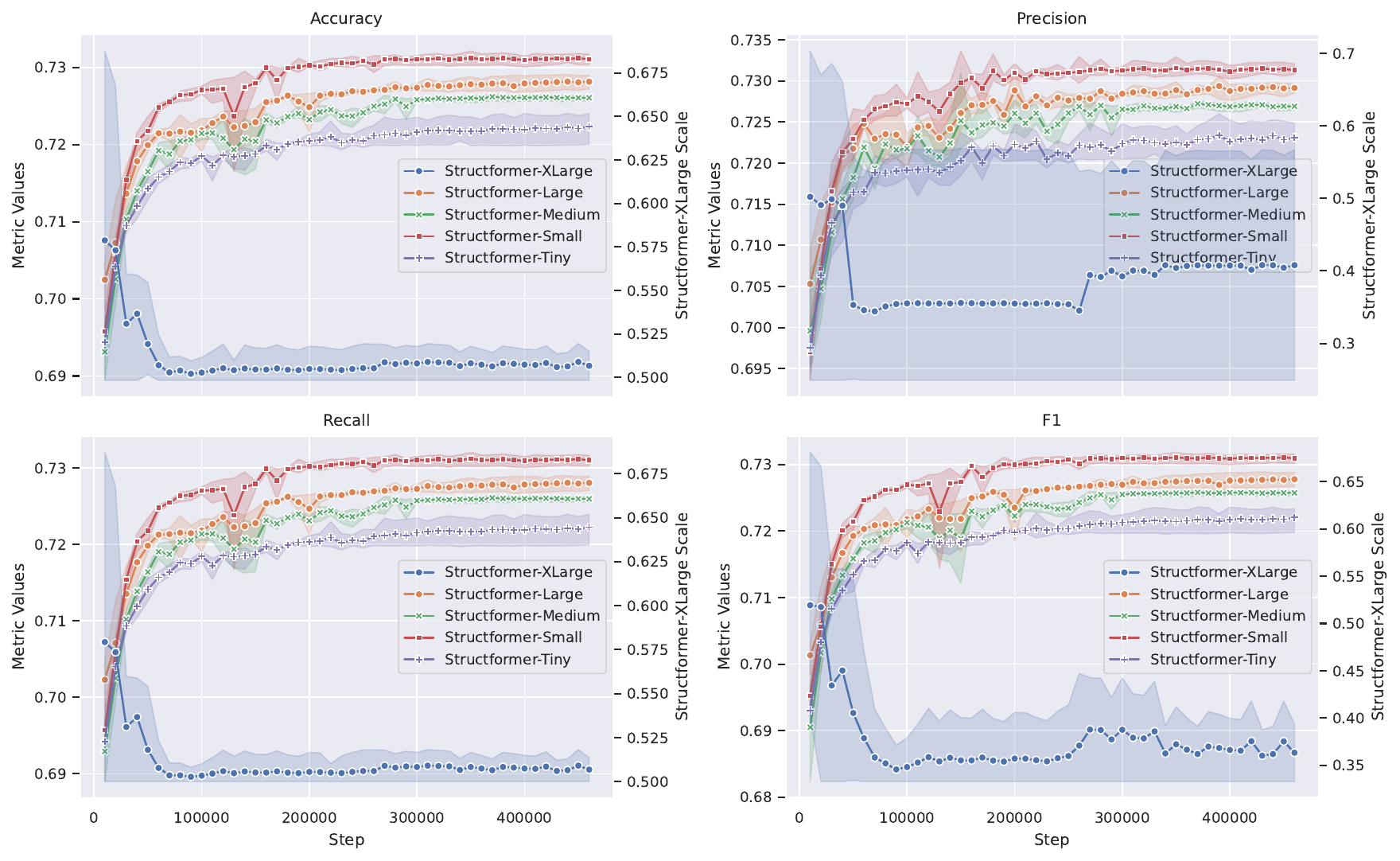}
\caption{Structformer binary classification metrics curves.}
\label{fig:binary_metrics_curves_double_axis}
\end{figure}

\begin{figure}[ht]
\centering
\includegraphics[width=\textwidth]{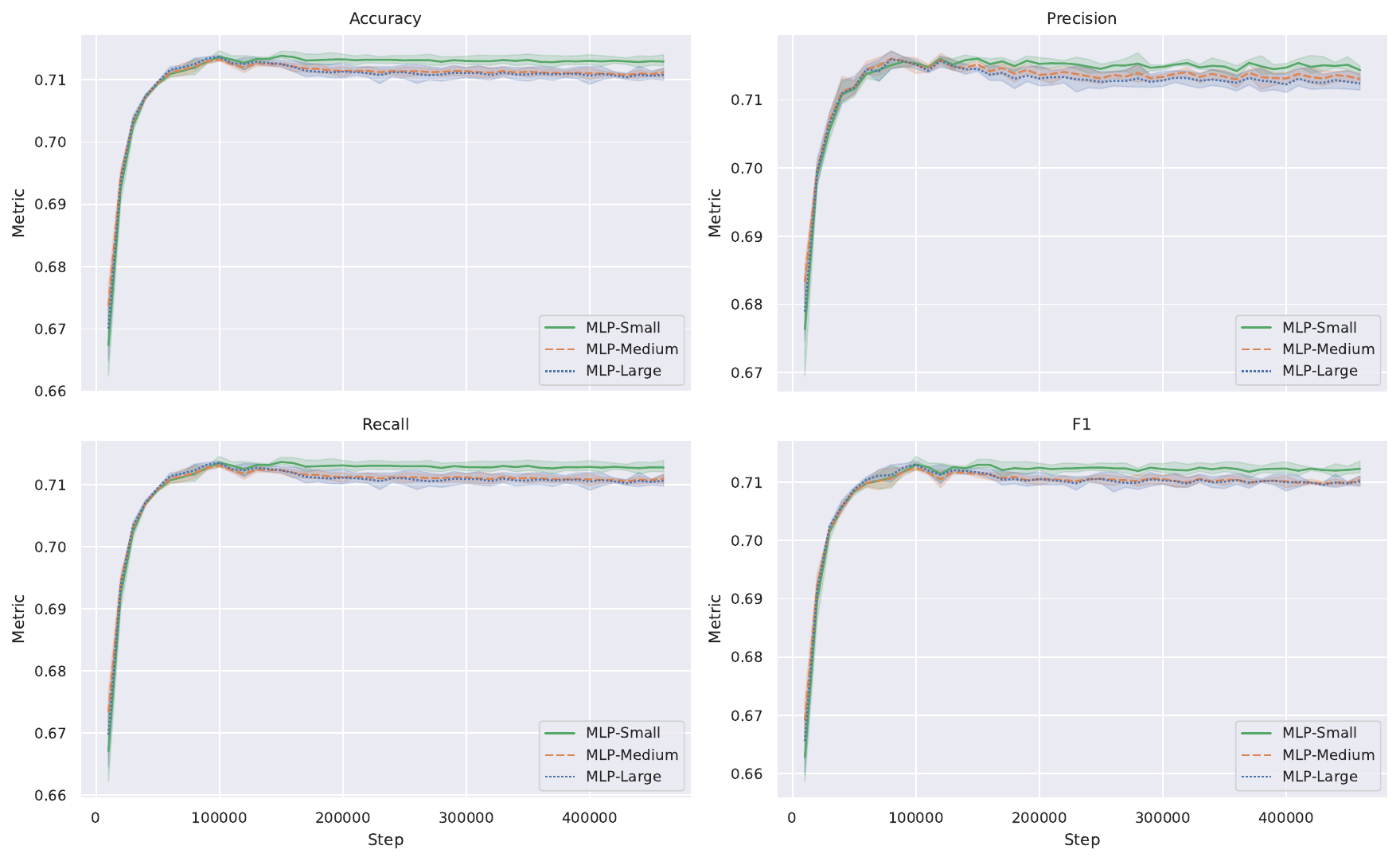}
\caption{MLP binary classification metrics curves.}
\label{fig:mlp_binary_metrics_curves}
\end{figure}

\begin{figure}[ht]
\centering
\includegraphics[width=\textwidth]{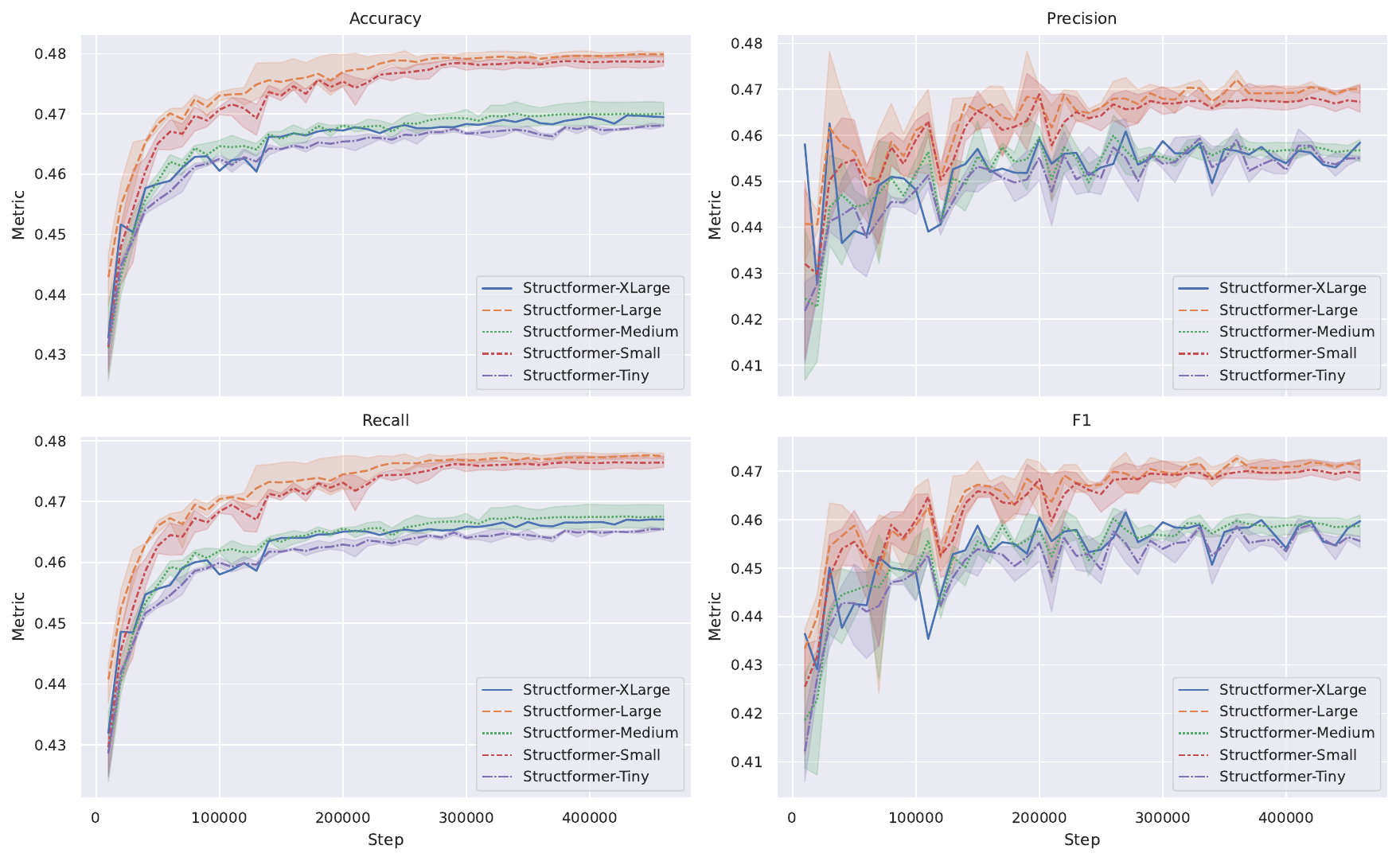}
\caption{Structformer multi-class classification metrics curves.}
\label{fig:multiclass_metrics_curves}
\end{figure}

\begin{figure}[ht]
\centering
\includegraphics[width=\textwidth]{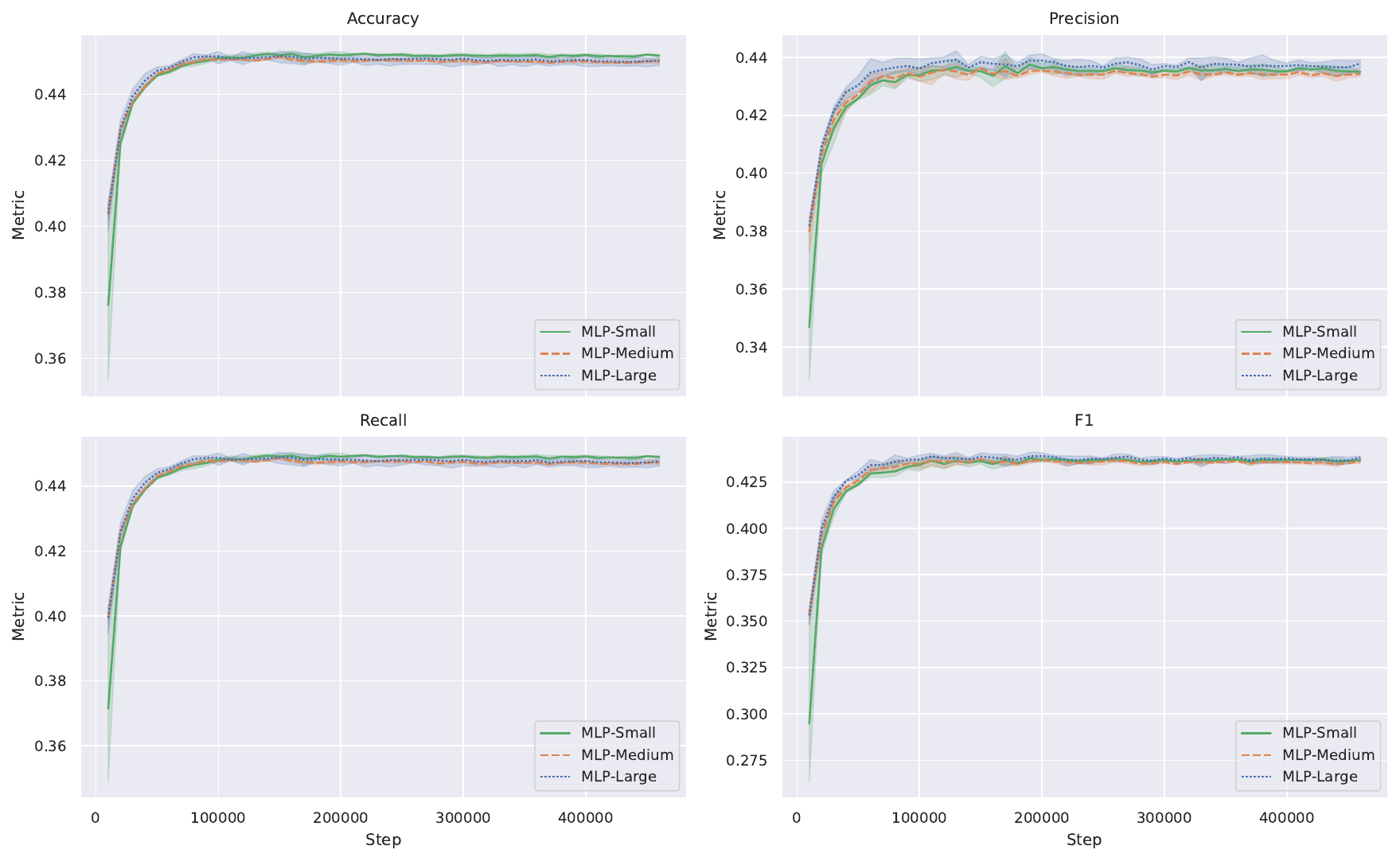}
\caption{MLP multi-class classification metrics curves.}
\label{fig:mlp_multiclass_metrics_curves}
\end{figure}

\end{document}